\pdfoutput=1

\documentclass[11pt]{article}

\usepackage{acl}

\usepackage{times}
\usepackage{latexsym}

\usepackage{amssymb}
\usepackage[dvipsnames]{xcolor}

\usepackage{tcolorbox}
\usepackage{kantlipsum}
\usepackage{setspace}

\usepackage[T1]{fontenc}

\usepackage[utf8]{inputenc}
\usepackage[T1]{fontenc}    
\usepackage{hyperref}       
\usepackage{url}            
\usepackage{booktabs}       
\usepackage{amsfonts}       
\usepackage{nicefrac}       
\usepackage{multirow}
\usepackage{multirow, makecell}
\usepackage{tabularx}

\newcommand\clearrow{\global\let\rowmac\relax}
\clearrow
\newcommand\blfootnote[1]{%
  \begingroup
  \renewcommand\thefootnote{}\footnote{#1}%
  \addtocounter{footnote}{-1}%
  \endgroup
}

\usepackage{booktabs,arydshln}
\makeatletter
\def\adl@drawiv#1#2#3{%
        \hskip.5\tabcolsep
        \xleaders#3{#2.5\@tempdimb #1{1}#2.5\@tempdimb}%
                #2\z@ plus1fil minus1fil\relax
        \hskip.5\tabcolsep}
\newcommand{\cdashlinelr}[1]{%
  \noalign{\vskip\aboverulesep
           \global\let\@dashdrawstore\adl@draw
           \global\let\adl@draw\adl@drawiv}
  \cdashline{#1}
  \noalign{\global\let\adl@draw\@dashdrawstore
           \vskip\belowrulesep}}
\makeatother

\usepackage{times}
\usepackage{latexsym}
\usepackage{soul,color}
\usepackage{todonotes} 
\usepackage[normalem]{ulem}
\usepackage{tikz}
\usepackage{array}
\usepackage[T1]{fontenc}
\usepackage{wrapfig}

\usepackage{fixltx2e}
\usepackage{amsmath}
\usepackage{textgreek}
\usepackage{enumitem,bbding,etoolbox,calc,pifont}

\usepackage{cleveref}

\usepackage{soul}

\usepackage{xspace,mfirstuc,tabulary}

\newcommand\Dreaddit{\textsc{Dreaddit}\xspace}

\usepackage{microtype}

\title{Monte Carlo Tree Search for Interpreting Stress in Natural Language}

\author{Kyle Swanson$^\star$, Joy Hsu$^\star$, Mirac Suzgun$^\star$ \\
  Department of Computer Science \\
  Stanford University \\
  \texttt{\{swansonk, joycj, msuzgun\}@stanford.edu} \\}

\begin{document}
\maketitle
\begin{abstract}
Natural language processing can facilitate the analysis of a person's mental state from text they have written. Previous studies have developed models that can predict whether a person is experiencing a mental health condition from social media posts with high accuracy. Yet, these models cannot explain \textit{why} the person is experiencing a particular mental state. In this work, we present a new method for explaining a person's mental state from text using Monte Carlo tree search (MCTS). Our MCTS algorithm employs trained classification models to guide the search for key phrases that explain the writer's mental state in a concise, interpretable manner. Furthermore, our algorithm can find both explanations that depend on the particular context of the text (e.g., a recent breakup) and those that are context-independent. Using a dataset of Reddit posts that exhibit stress, we demonstrate the ability of our MCTS algorithm to identify interpretable explanations for a person's feeling of stress in both a context-dependent and context-independent manner.\blfootnote{$^\star$Denotes equal contribution.}\footnote{Code and models are available at \url{https://github.com/swansonk14/MCTS_Interpretability}.}
\end{abstract}

\section{Introduction}

Disabilities associated with mental health conditions pose a significant challenge for many people around the world \cite{stauder2010worldwide, de2013social, chen2018mood}. To help people suffering from these conditions, it is crucial to identify those who are experiencing a mental health condition and understand the underlying causes.

Natural language processing (NLP) can help by analyzing a person's mental state based on the text they have written. Previous studies \cite{turcan2019dreaddit, demszky2020goemotions, gjurkovic2020pandora, ansari2021data} have demonstrated the ability of NLP models to process social media posts and predict stress, depression, and a range of emotions. These methods, however, are not able to explain \textit{why} the person might be feeling the way they are, even if that information is clearly contained in the text analyzed by the model.

In this work, we seek to explain the underlying causes of a person's mental state from their writing. We formulate such an explanation as a small set of phrases from the text that is sufficient to explain the person's mental state. We wish to identify two complementary types of explanations: those that are particular to the situation the person is in, which we call \emph{context-dependent}, and those that could appear across different contexts, which we call \emph{context-independent}. Figure \ref{fig:explanation_example} shows an illustrating example. Identifying both types of explanations not only enhances our understanding of the underlying sources of a person's mental state but also provides insights into how one's mental state can be affected by general and specific causes.

\begin{figure}[t]
    \begin{tcolorbox}[fontupper=\footnotesize]
        \textbf{r/Relationships:} I can't believe this. \textcolor{blue}{My boyfriend just cheated on me} and then he bragged about it on twitter. What kind of a messed up person would do that? \textcolor{red}{I'm so angry with him} and I'm sure we're going to \textcolor{red}{have a huge fight about this} when I see him tomorrow.
    \end{tcolorbox}
    \vspace{-1em}
    \caption{A fictitious example of text exhibiting stress in the relationships context and two explanations for that stress. The explanation in \textcolor{blue}{blue} is \emph{context-dependent} (specific to relationships) while the explanation in \textcolor{red}{red} is \emph{context-independent} (general to any disagreement).}
    \label{fig:explanation_example}
\end{figure}

To this end, we develop a novel Monte Carlo tree search (MCTS) algorithm that can effectively identify explanations that are either context-dependent or context-independent by leveraging the semantic capabilities of trained NLP models. We, both quantitatively and qualitatively, demonstrate the efficacy of this approach to explain a person's mental state using a dataset of Reddit posts that exhibit stress \cite{turcan2019dreaddit}.

\section{Related Work}
\textbf{Mental Health Prediction.} Previous studies have tackled the task of mental health disability classification, using methods ranging from classical supervised techniques such as SVMs, logistic regression, Naive Bayes, MLPs, and decision trees to deeper models such as CNNs and GRUs \cite{turcan2019dreaddit, gjurkovic2020pandora, ansari2021data, depsign-acl}. Other approaches utilize pre-trained, large language models with fine-tuning on specific mental health datasets \cite{ji2021mentalbert, matovsevic2021stressformers, mauriello2021sad}, which takes advantage of models trained on significantly larger datasets to speed up training and increase accuracy. \citet{turcan2019dreaddit} specifically focus on the task of stress prediction in Reddit posts, and they show that large BERT-based models outperform smaller models such as CNNs and logistic regression.

\textbf{NLP Explainability.} Explainability in NLP is an emerging topic of interest as language models have become larger and more accurate at the expense of reduced interpretability. Common methods for explainability include feature importance reporting across lexical or latent features \cite{danilevsky2020survey}, model-agnostic approaches that extract post-hoc explanations \cite{ribeiro2016model}, and analogy-based explanations \cite{croce2019auditing}. Prior works have also focused on rationale identification \cite{lei2016rationalizing} and text matching rationalization \cite{swanson2020rationalizing}, where models are designed to select small, interpretable segments of text when making predictions. Attention has also been used as a form of interpretability, but attention weights do not always correlate with impact on the model's prediction, potentially limiting their usefulness \cite{serrano2019attention}. In this work, we propose to use Monte Carlo tree search \cite{silver2016mastering, chaudhry2018feature, jin2020multi, albrecht2021interpretable, pmlr-v139-yuan21c} as a post-hoc explainability method that can be applied to any model to flexibly identify multiple types of explanations for a model's predictions.

\section{The \Dreaddit Dataset}
The \Dreaddit dataset \cite{turcan2019dreaddit} contains 3,553 Reddit posts that have human-annotated binary stress labels denoting whether a given text contains evidence of stress. Each post belongs to one of ten subreddits (e.g., ``r/Relationships''), which we consider to be the context of the post. The posts are split into 2,838 train posts and 715 test posts. Figures \ref{fig:stress_label_distribution} and \ref{fig:subreddit_distribution} (see Appendix) show the distributions of the stress labels and subreddit categories for the train and test sets.

\section{Method}
\begin{figure*}
    \centering
    \includegraphics[width=1.0\textwidth]{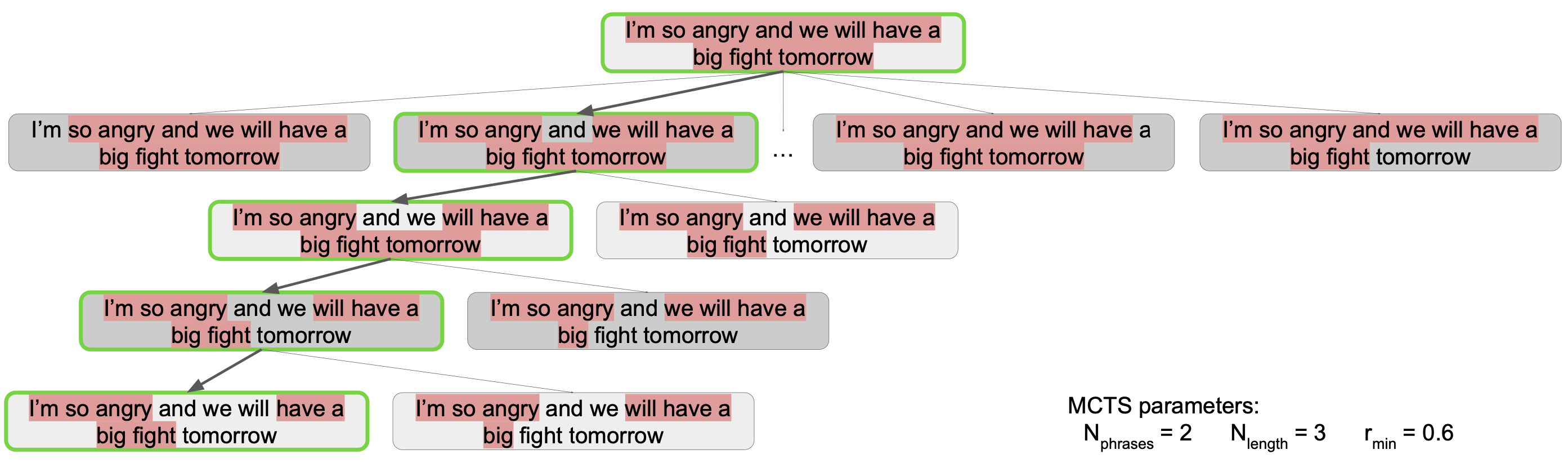}
    \caption{A portion of the tree of explanations searched by MCTS for an example text. Red indicates the text that is currently included in the explanation. The root of the tree is an explanation with a single phrase containing all the text. Each node in the tree can be expanded by removing the first or last token of a phrase or by removing a token in the middle of the phrase (constrained by certain MCTS parameters). Once a minimum number of tokens has been reached, the resulting explanation is given a reward based on the predictions of the stress and context models.}
    \label{fig:mcts}
\end{figure*}

We assume that we have access to a training corpus $\mathcal{D}_{\text{train}}$ and a test corpus $\mathcal{D}_{\text{test}}$ to train and evaluate our models, respectively. The training corpus, $\mathcal{D}_{\text{train}}=(\mathbf{t}_i, \mathbf{s}_i, \mathbf{c}_i)_{i\in [1, n]}$, is a set of tuples, where each tuple contains a text $\mathbf{t}_i=\{t_i^1, \cdots, t_i^{l_i}\} \in \mathbf{T}$ consisting of $l_i$ tokens, its corresponding stress indicator $\mathbf{s}_i \in \mathbf{S}=\{0,1\}$ denoting whether $\mathbf{t}_i$ contains evidence of stress, and a context label $\mathbf{c}_i \in \mathbf{C}$ indicating the subreddit category the text belongs to. Similarly, we assume $\mathcal{D}_{\text{test}}=(\mathbf{t}_i, \mathbf{s}_i, \mathbf{c}_i)_{i\in [1, m]}$.

\subsection{Classification of Stress and Context}

We consider two types of classification tasks, namely binary stress classification and multi-class context (subreddit) classification. 
We refer to a model trained for the former task as a \texttt{stress} classifier, which can be thought of as a function mapping a piece of text $\mathbf{t} \in \mathbf{T}$ to a likelihood $p \in [0, 1]$. We refer to a model trained for the latter as a \texttt{context} classifier, which can be thought of as a function mapping a piece of text $\mathbf{t} \in \mathbf{T}$ to a probability simplex $\triangle^{|\mathbf{C}|-1}$.

We build simple stress and context prediction models using Bernoulli and Multinomial Naive Bayes, Support Vector Machine \cite{Platt99probabilisticoutputs}, and Multilayer Perceptron \cite{hinton_mlp}. All of these models use vectors of word counts\footnote{We use \texttt{CountVectorizer} from \texttt{scikit-learn} fit on the training set with all default parameters.} as inputs. We also build large BERT-based models by adding a classification layer on top of the MentalRoBERTa model of \citet{ji2021mentalbert} and then fine-tuning the model on the training set.

\subsection{Definition of an Explanation}

An interpretable explanation for a person's stress should consist of a small set of phrases from the full text that captures the core reasons behind the stress discussed within the text.

Formally, for a given piece of text in the corpus $\mathbf{t} \in \mathbf{T}$ that is labeled as stressed ($\mathbf{s}=1$), we define an \emph{explanation} as a set of phrases $\mathbf{E} = \{\mathbf{p}_1, \mathbf{p}_2, \dots, \mathbf{p}_k\}$ where each phrase $\mathbf{p}_j$ is a set of $n_j$ contiguous tokens in the text, that is, $\mathbf{p}_j = \{t_{l}, t_{l+1}, \dots, t_{l + {n_j} - 1}\}$ for some $l \in \{1, 2, \dots, |\mathbf{t}| - n_j + 1\}$. Furthermore, the phrases must be non-overlapping, which means that $\mathbf{p}_j \cap \mathbf{p}_{j'} = \emptyset \quad \forall j \neq j' \in \{1, 2, \dots, |\mathbf{E}|\}$. In order to ensure interpretability, the explanation $\mathbf{E}$ must satisfy three conditions.

\textbf{a. Phrase count:} $|\mathbf{E}| \leq N_{phrases}$, meaning the explanation must contain at most $N_{\textrm{phrases}}$ phrases. Too many phrases would impede interpretability.
    
\textbf{b. Phrase length:} $|\mathbf{p_j}| \geq N_{length} \quad \forall j \in \{1, 2, \dots, |\mathbf{E}|\}$, meaning each phrase must have at least $N_{length}$ tokens, preventing phrases that are too short to carry any meaning.
    
\textbf{c. Proportion of tokens:} $r_{min} \leq r(\mathbf{E}) \leq r_{max}$ where $r(\mathbf{E}) = \frac{1}{|\mathbf{t}|} \sum_{j=1}^{|\mathbf{E}|} |\mathbf{p_j}|$ is the proportion of tokens in the text that are included in the explanation and $0 \leq r_{min} \leq r_{max} \leq 1$ are lower and upper bounds on the proportion of tokens in the explanation. This constrains the overall verbosity of the explanation to a reasonable range.

\subsection{Context-Dependent and Independent Explanations of Stress}

We are interested in identifying two specific types of explanations for stress: one that depends on the context of the text and one that is independent of that context. We will refer to the context-dependent explanation as $\mathbf{E}_{dep}$ and to the context-independent explanation as $\mathbf{E}_{ind}$.

In both cases, since the explanation must explain the stress in the text, the stress must be evident from just the text contained in the phrases of the explanation. We can verify this by using our stress classification model. Specifically, we want an explanation such that the average stress prediction across the phrases of the explanation is close to $1$. Hence for both $\mathbf{E}_{dep}$ and $\mathbf{E}_{ind}$, we want
\vspace{-1.3mm}
$$S(\mathbf{E}) = \frac{1}{|\mathbf{E}|} \sum_{j=1}^{|\mathbf{E}|} \texttt{stress}(\mathbf{p}_j) \approx 1$$

\noindent
where $S(\mathbf{E})$ is the average stress across the phrases of the explanation.

However, the phrases of the context-dependent explanation $\mathbf{E}_{dep}$ should indicate the context of the text while the context-independent explanation $\mathbf{E}_{ind}$ should not. We enforce this by examining the entropy of the predictions of our context classification model. If the phrases of an explanation have low entropy, then the model is relatively sure of the context; hence, that explanation is context-dependent. If the entropy is high, then the model is unsure of the context and the explanation is context-independent. Formally, if we define
\vspace{-1.3mm}
$$H(\mathbf{E}) = \frac{1}{|\mathbf{E}|} \sum_{j=1}^{|\mathbf{E}|} \texttt{entropy}(\texttt{context}(\mathbf{p}_j))$$

\noindent
as the average Shannon entropy of the context predictions across phrases, we want $H(\mathbf{E_{dep}}) \approx 0$ and $H(\mathbf{E_{ind}}) \approx e_{\textrm{max}}$ where $e_{\textrm{max}}$ is the maximum entropy (viz., entropy of a uniform distribution over contexts).

\subsection{Finding Explanations with MCTS}

We use the MCTS framework established in \citet{silver2016mastering}, but we modify the search tree and the reward function to suite our purposes (see Figure~\ref{fig:mcts}). Each node in the tree represents an explanation $\mathbf{E} = \{\mathbf{p}_1, \mathbf{p}_2, \dots, \mathbf{p}_k\}$. The root of the tree represents the whole text piece as a single phrase, i.e., $\mathbf{E}_{root} = \{\mathbf{t}\}$. When the search is at a given node in the tree, there are two options for expanding the next node: (i) remove the first or last token in any phrase, as long as the shortened phrase still contains at least $N_{length}$ tokens, or (ii) remove a token in the middle of a phrase, thus breaking it into two phrases, as long as both resulting phrases have at least $N_{length}$ tokens and the total number of phrases does not exceed $N_{phrases}$.

The search continues to expand nodes in the tree until either the current node cannot be expanded using either of the two rules above or the explanation at the current node contains too few tokens, i.e., $r(\mathbf{E}) \leq r_{min}$. This node serves as a leaf node and is given a reward equal to
\vspace{-1.3mm}
$$R(\mathbf{E}) = S(\mathbf{E}) + I \cdot \alpha \cdot H(\mathbf{E})$$

\noindent for some $I \in \{-1, +1\}$ and $\alpha \geq 0$. We use $I = +1$ to select for high entropy (context-independent) explanations and $I = -1$ to select for low entropy (context-dependent) explanations. This reward is propagated back to all the nodes on the path from the root to this leaf node according to the update rules from \citet{silver2016mastering}. After the search is complete, the best explanation $\mathbf{\hat{E}}$ is selected as
\vspace{-1.3mm}
$$\mathbf{\hat{E}} = \underset{\mathbf{E}}{\textrm{argmax}}\ R(\mathbf{E})\ \ \textrm{s.t.}\ r(\mathbf{E}) \leq r_{max},$$

\noindent which means $\mathbf{\hat{E}}$ is the explanation in the search tree that maximizes the reward while satisfying the condition on the maximum proportion of tokens. The other interpretability conditions are guaranteed by the rules of the search tree expansion.

\begin{table}[t]
\small 
\scalebox{0.90}{
\begin{tabular}{c | c  c  c  c }
\toprule
\bf{Model} & \bf{Precision} & \bf{Recall} &\bf{F-1} & \bf{Accuracy} \\ \toprule
Bernoulli NB & 0.69 & 0.84 & 0.75 & 0.72 \\
Multinomial NB & 0.68 & 0.87 & 0.76 & 0.72 \\
SVM & 0.71 & 0.77 & 0.74 & 0.72 \\
MLP & 0.71 & 0.74 & 0.73 & 0.71 \\
MentalRoBERTa\textsuperscript{FT} & 0.78 & 0.90 & 0.84 & 0.82 \\
\bottomrule
\end{tabular}
}
\caption{Performances of stress classifiers on the test set of \Dreaddit. While non-neural classifiers could not surpass 72\% accuracy, the MentalRoBERTa\textsuperscript{FT} model fine-tuned on the \Dreaddit train set yielded 82\% accuracy. Here, the superscript \textsuperscript{FT} denotes that the model was fine-tuned.}

\label{tab:StressClassificationResults}
\end{table}

\section{Experiments}
All of our experiments were run on the \Dreaddit dataset. We report results of our stress and context classification models and share findings of our MCTS explanation algorithm.

\begin{table}[t]
\small 
\scalebox{0.90}{
\begin{tabular}{c | c  c  c  c }
\toprule
\bf{Model} & \bf{Precision} & \bf{Recall} &\bf{F-1} & \bf{Accuracy} \\ \toprule
Bernoulli NB & 0.81 & 0.75 & 0.76 & 0.80 \\
Multinomial NB & 0.77 & 0.75 & 0.75 & 0.79 \\
SVM & 0.76 & 0.72 & 0.74 & 0.76 \\
MLP & 0.78 & 0.78 & 0.78 & 0.79 \\
MentalRoBERTa\textsuperscript{FT} & 0.85 & 0.86 & 0.86 & 0.87 \\
\bottomrule
\end{tabular}
}
\caption{Performances of context classifiers. We restricted our focus to three subreddits: ``anxiety,'' ``assistance,'' ``relationships.'' The fine-tuned MentalRoBERTa\textsuperscript{FT} model yielded the best results with 87\% accuracy.}
\label{tab:ContextClassificationResults}
\end{table}

\subsection{Classification}
\label{exp_classification}

As Table~\ref{tab:StressClassificationResults} illustrates, basic stress classification models, such as Naive Bayes classifiers, SVMs, and MLPs, performed reasonably on the test set of \Dreaddit. The MentalRoBERTa\textsuperscript{FT} model for stress fine-tuned on the training set of \Dreaddit for five epochs, however, was able to outperform all the other models, achieving an accuracy score of 82\% and demonstrating the efficacy of the pre-training on mental health data\footnote{In contrast, the RoBERTa model trained from scratch achieved an accuracy score of almost 80\%.}. Our results on the stress classification task are consistent with those of \citet{turcan2019dreaddit}. Table~\ref{tab:ContextClassificationResults} reports the performance of various models on the multi-class subreddit category classification. Here, we limited our attention to three categories, namely ``anxiety,'' ``assistance,'' and ``relationships.'' The Reddit posts in these categories embody various distinct everyday, financial, and interpersonal stress factors, but at the same time, they seem to have common (context-independent) stress elements. In this context classification task, all models were able to go beyond the 75\% accuracy level, but MentalRoBERTa\textsuperscript{FT} yielded the highest accuracy.

\subsection{Explainability}
\label{exp_explainability}
We demonstrate our MCTS approach to explainability using the same three categories as above. We use stress and context classification models implemented with Multinomial NB, MLP, and MentalRoBERTa\textsuperscript{FT}. For each of these models, we apply MCTS to identify explanations for each of the 166 test texts that is labeled as stressed and belongs to one of our three categories. We use the interpretability conditions $N_{phrases} = 3$, $N_{length} = 5$, $r_{min} = 0.2$, and $r_{max} = 0.5$ for all experiments\footnote{These choices are arbitrary and could easily be changed.}, and we use $\alpha = 10$ except where otherwise noted.

\begin{table}[t]
\centering \small 
\centering
\scalebox{0.90}{\begin{tabular}{llccc}
\toprule
 & & \textbf{Original} & \textbf{Dependent} & \textbf{Independent} \\
\toprule
\multirowcell{2}{\textbf{MNB}} & S & 0.850 $\pm$ 0.317 & 0.706 $\pm$ 0.190 & 0.617 $\pm$ 0.124 \\
& E & 0.047 $\pm$ 0.140 & 0.274 $\pm$ 0.181 & 0.942 $\pm$ 0.086 \\  \midrule
\multirowcell{2}{\textbf{MLP}} & S & 0.725 $\pm$ 0.383 & 0.512 $\pm$ 0.194 & 0.546 $\pm$ 0.145 \\
& E & 0.214 $\pm$ 0.274 & 0.766 $\pm$ 0.163 & 1.067 $\pm$ 0.022 \\  \midrule
\multirowcell{2}{\textbf{MRB}} & S & 0.878 $\pm$ 0.324 & 0.830 $\pm$ 0.220 & 0.430 $\pm$ 0.273 \\
& E & 0.042 $\pm$ 0.124 & 0.019 $\pm$ 0.018 & 0.640 $\pm$ 0.171 \\  
\bottomrule
\end{tabular}}
\caption{Stress (S) and context entropy (E) for original text, context-dependent explanation, and context-independent explanation for the Multinomial Naive Bayes (MNB), Multilayer Perceptron (MLP), and Mental RoBERTa (MRB) models. Results were generated through MCTS with stress and context entropy averaged over the test set. The Wilcoxon signed rank test \cite{wilcoxon} between dependent and independent entropy is $p < 0.0001$ for all models, indicating a very significant difference as desired.}
\label{results_explainability}
\end{table}

\begin{figure}[t]
\centering
{\includegraphics[width=\columnwidth]{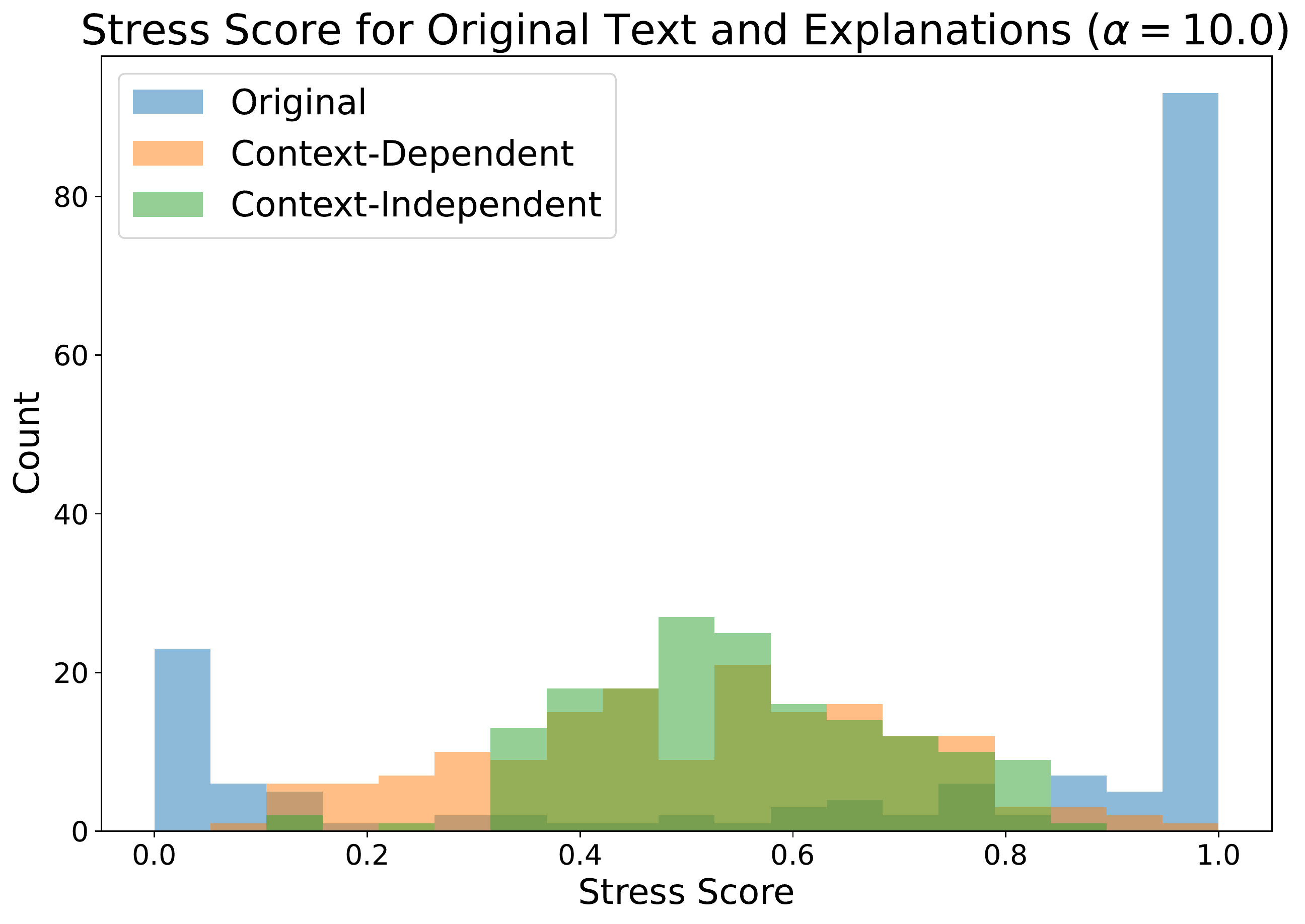}}
\caption{Histogram of stress scores for the original text and for the context-dependent and context-independent explanations extracted by our MCTS algorithm using an MLP model. Although stress is often higher in the original text than in the extracted explanations, the explanations still maintain a meaningful amount of stress.}
\label{fig:stress_explanations}
\end{figure}

\begin{figure}[t]
\centering
{\includegraphics[width=\columnwidth]{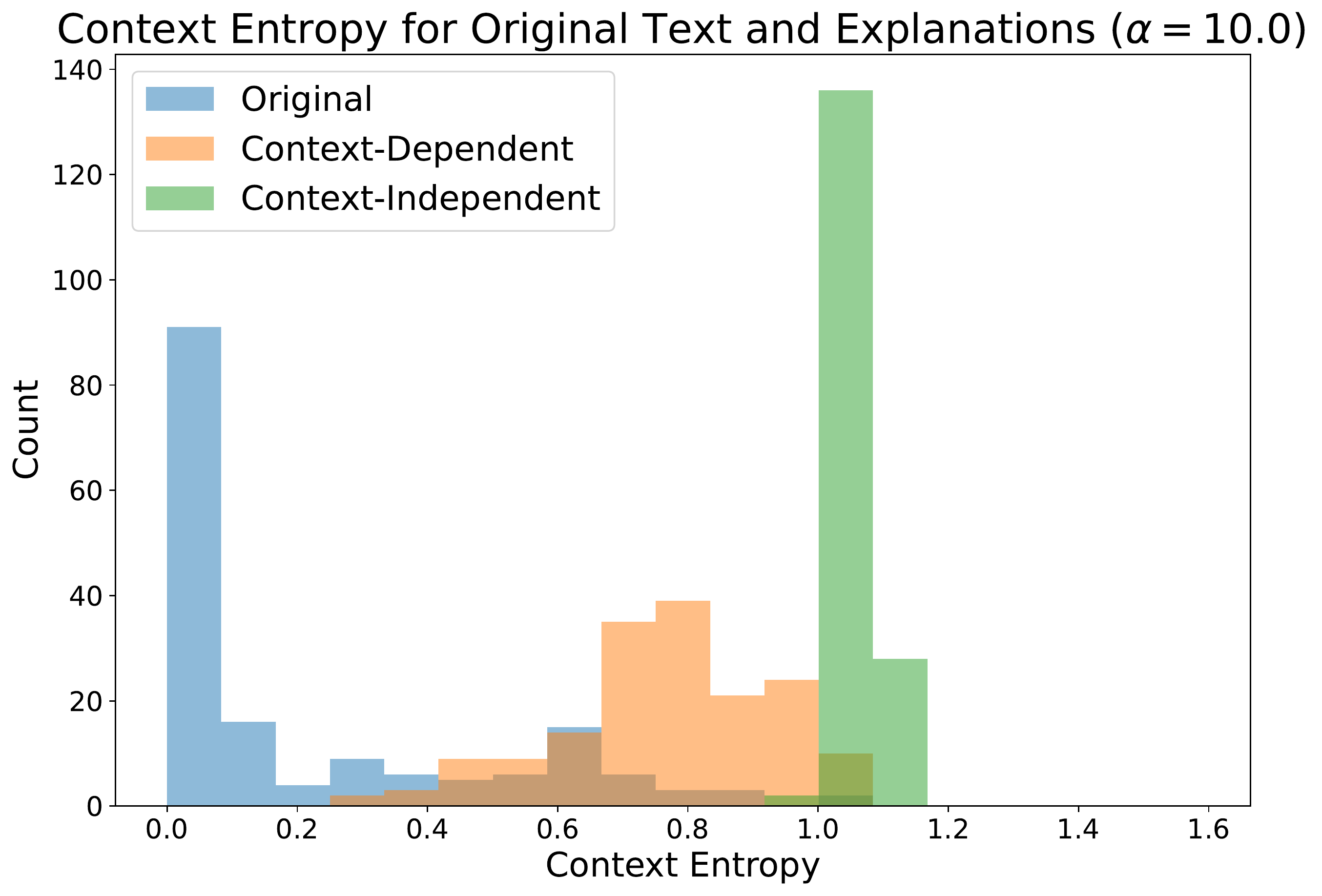}}
\caption{Histogram of context entropy for the original text and for the context-dependent and context-independent explanations extracted by our MCTS algorithm using an MLP model. The context-independent explanations clearly have much higher context entropy than the context-dependent explanations as desired.}
\label{fig:entropy_explanations}
\end{figure}

We quantitatively evaluate the explanations produced by MCTS. In Table~\ref{results_explainability}, we show the average stress and context entropy scores of the original text and of the context-dependent and context-independent explanations. Our method is able to maintain a reasonably high and consistent level of stress across the explanations while modulating the context entropy appropriately for the two different types of explanations. This indicates that our approach can identify both context-dependent and context-independent sources of stress.

Figures \ref{fig:stress_explanations} and \ref{fig:entropy_explanations} further illustrate this result for the MLP model by showing the full distribution of stress and context entropy scores across the test examples. Figures \ref{fig:MNB}, \ref{fig:MLP}, and \ref{fig:BERT} in the Appendix show the stress and context entropy distributions for all three models and for different values of $\alpha$. Lower $\alpha$ increases stress but decreases the difference in entropy between the two types of explanations while higher $\alpha$ decreases stress but increases the difference in entropy. This shows the flexibility of MCTS to select different types of explanations without retraining the classifiers.

Furthermore, we qualitatively demonstrate our approach. Tables \ref{tab:results_explainability_qualitative_anxiety}, \ref{tab:results_explainability_qualitative_assistance}, and \ref{tab:results_explainability_qualitative_relationships} in the Appendix show examples from each of the three subreddits that illustrate how our method captures different underlying sources of stress in an interpretable manner.

\section{Conclusion}
\vspace{-0.5em}
We propose a novel interpretability method for explaining stress in context-dependent and independent manners using Monte Carlo tree search. We demonstrate the effectiveness of our method by extracting both types of explanations from Reddit posts that exhibit stress. Although this work focuses on stress, our MCTS-based explanation framework is extremely flexible and can be applied to a wide variety of NLP models and prediction problems simply by specifying the appropriate reward function and interpretability conditions for the search tree. As in our work, the reward function can include multiple objectives with different weights, making it possible to extract a variety of explanations for added interpretability. Future work should further explore the range of explanations enabled by our framework. We hope that our explanation framework can improve understanding of the root causes of mental health conditions as well as provide interpretability for a variety of NLP tasks.

\newpage
\section*{Acknowledgements}
We would like to thank Margalit Glasgow, Masha Karelina, Megha Patel, Biscuit Russell, and Tayfun M. H. Mezarci for helpful comments and discussions. Swanson and Hsu gratefully acknowledge the support of the Knight-Hennessy Scholarship, Hsu gratefully acknowledges the support of the NSF GRFP, and Suzgun gratefully acknowledges the support of a Johann, Thales, Williams \& Co. Graduate Fellowship. The authors also thank Dan Jurafsky for his support. The experiments presented in this paper were run on the Stanford NLP Cluster. Any opinions, findings, and conclusions expressed in this material are those of the authors and do not necessarily reflect the views of Stanford University. All errors remain our own.

\bibliography{custom}
\bibliographystyle{acl_natbib}

\appendix
\onecolumn

\section{Appendix}
\label{sec:appendix}

\subsection{Additional Stress and Context Entropy Results}

\begin{figure}[!h]
    \centering
    \includegraphics[width=\textwidth]{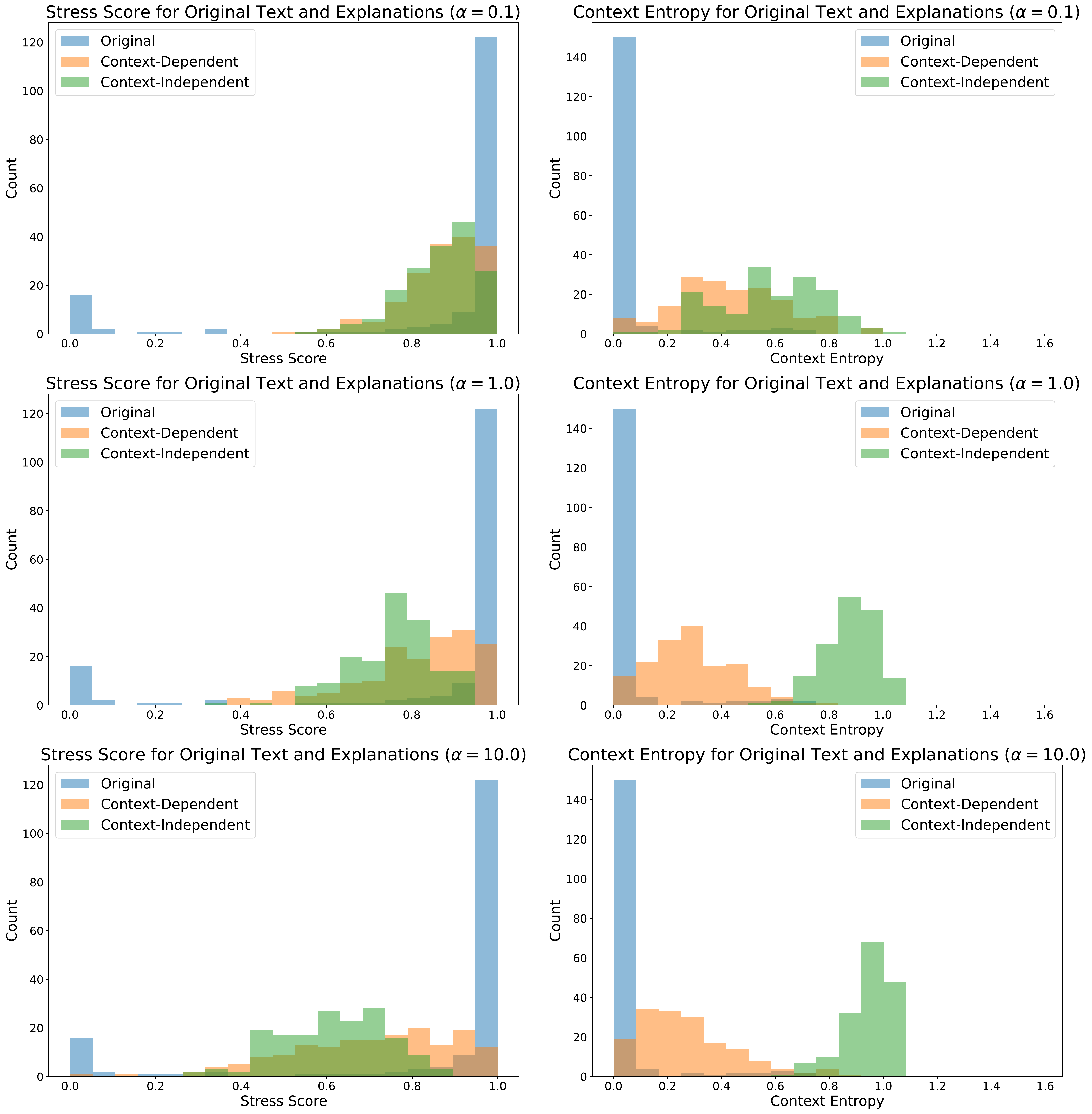}
    \caption{Histograms of stress and context entropy scores from the Multinomial Naive Bayes model for the original text and for the context-dependent and context-independent explanations extracted by our MCTS algorithm. The left column shows stress scores while the right column shows context entropy scores. From top to bottom, the rows show $\alpha = 0.1$, $\alpha = 1$, and $\alpha = 10$, where $\alpha$ controls the balance between stress and context entropy in the MCTS reward function. Higher $\alpha$ places less emphasis on stress and more emphasis on context entropy, resulting in a greater difference between context-dependent and context-independent entropy scores at the cost of lower stress.}
    \label{fig:MNB}
\end{figure}

\begin{figure}[!h]
    \centering
    \includegraphics[width=\textwidth]{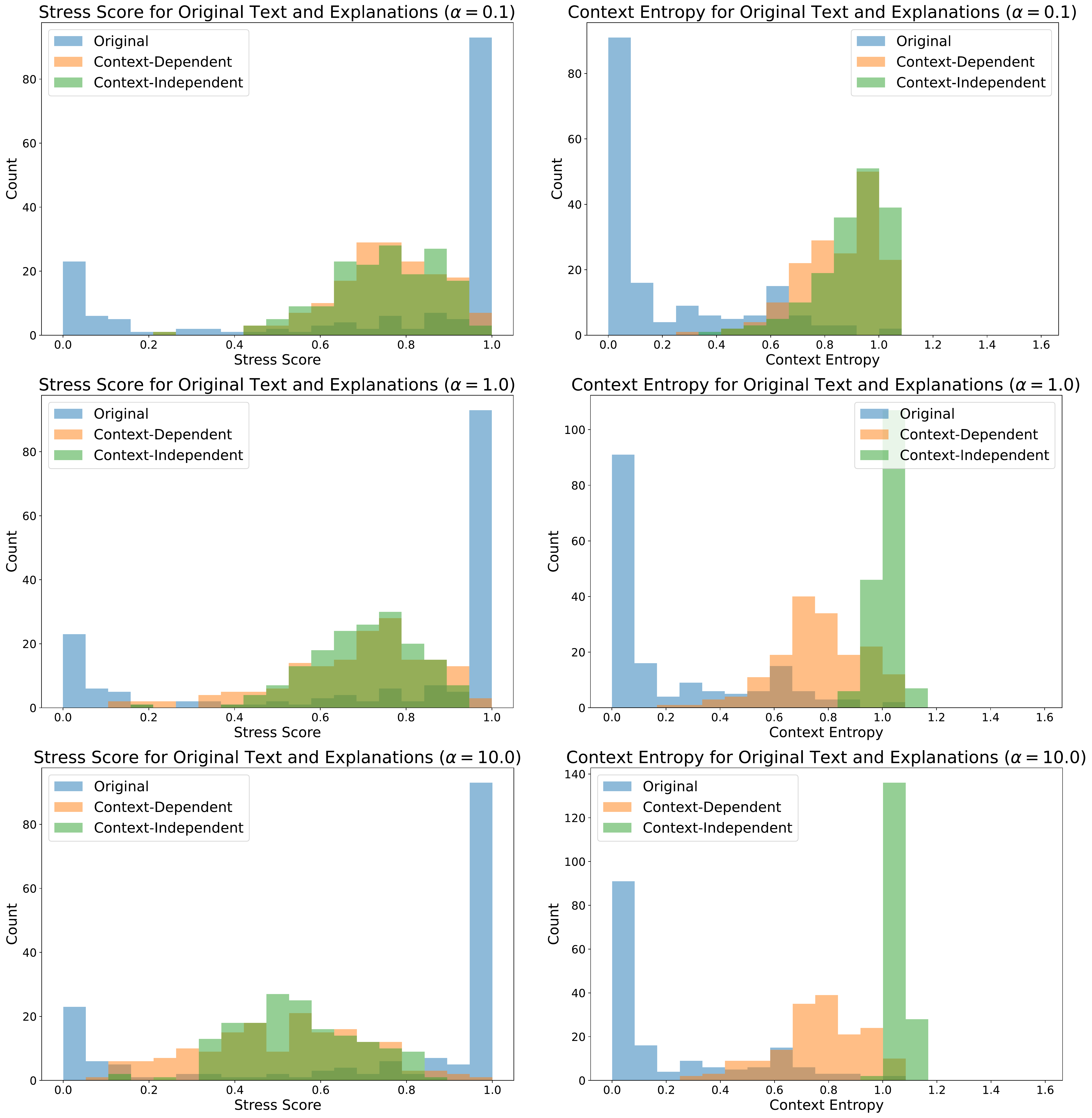}
    \caption{Histograms of stress and context entropy scores from the Multilayer Perceptron model for the original text and for the context-dependent and context-independent explanations extracted by our MCTS algorithm. The left column shows stress scores while the right column shows context entropy scores. From top to bottom, the rows show $\alpha = 0.1$, $\alpha = 1$, and $\alpha = 10$, where $\alpha$ controls the balance between stress and context entropy in the MCTS reward function. Higher $\alpha$ places less emphasis on stress and more emphasis on context entropy, resulting in a greater difference between context-dependent and context-independent entropy scores at the cost of lower stress.}
    \label{fig:MLP}
\end{figure}

\begin{figure}[!h]
    \centering
    \includegraphics[width=\textwidth]{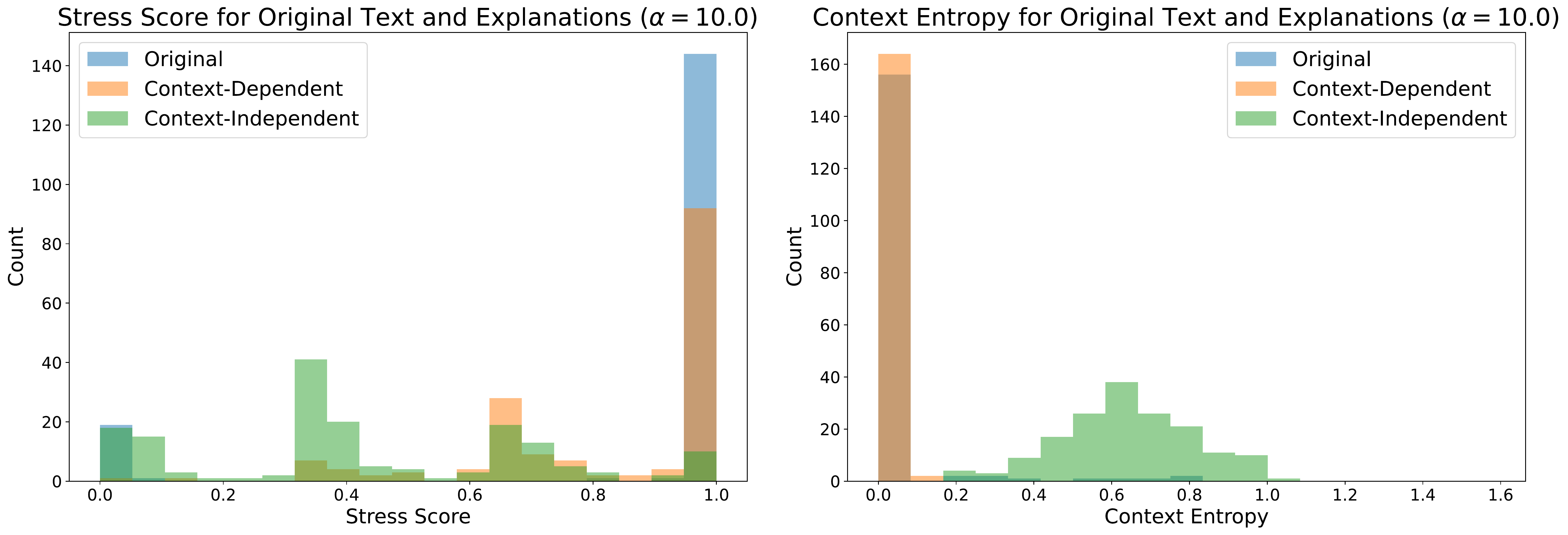}
    \caption{Histograms of stress and context entropy scores from the MentalRoBERTa\textsuperscript{FT} model for the original text and for the context-dependent and context-independent explanations extracted by our MCTS algorithm. The left plot shows stress scores while the right plot shows context entropy scores, both for $\alpha = 10$. Interestingly, the distributions are somewhat different from those of the Multinomial Naive Bayes (Figure \ref{fig:MNB}) and Multilayer Perceptron (Figure \ref{fig:MLP}) models. MentalRoBERTa\textsuperscript{FT} is capable of selecting different context-dependent and context-independent explanations as measured by entropy, but the model generally assigns more stress to context-dependent explanations than context-independent explanations, perhaps hinting at a meaningful difference between the types of explanations in terms of stress content.}
    \label{fig:BERT}
\end{figure}

Figures \ref{fig:MNB}, \ref{fig:MLP}, and \ref{fig:BERT} show the stress and context entropy distributions of the original text and the context-dependent and context-independent explanations across the 166 stressed test examples in the ``anxiety,'' ``assistance,'' and ``relationships'' subreddits for the Multinomial Naive Bayes, Multilayer Perceptron, and MentalRoBERTa\textsuperscript{FT} models, respectively. For the Multinomial Naive Bayes and Multilayer Perceptron models, we experimented with $\alpha \in \{0.1, 1, 10\}$, with higher $\alpha$ weighting context entropy more than stress in the MCTS reward function. For the MentalRoBERTa\textsuperscript{FT} model, we used $\alpha = 10$.

\newpage

\newpage

\subsection{Data Distribution}
\begin{figure}[!h]
\centering
{\includegraphics[width=0.70\columnwidth]{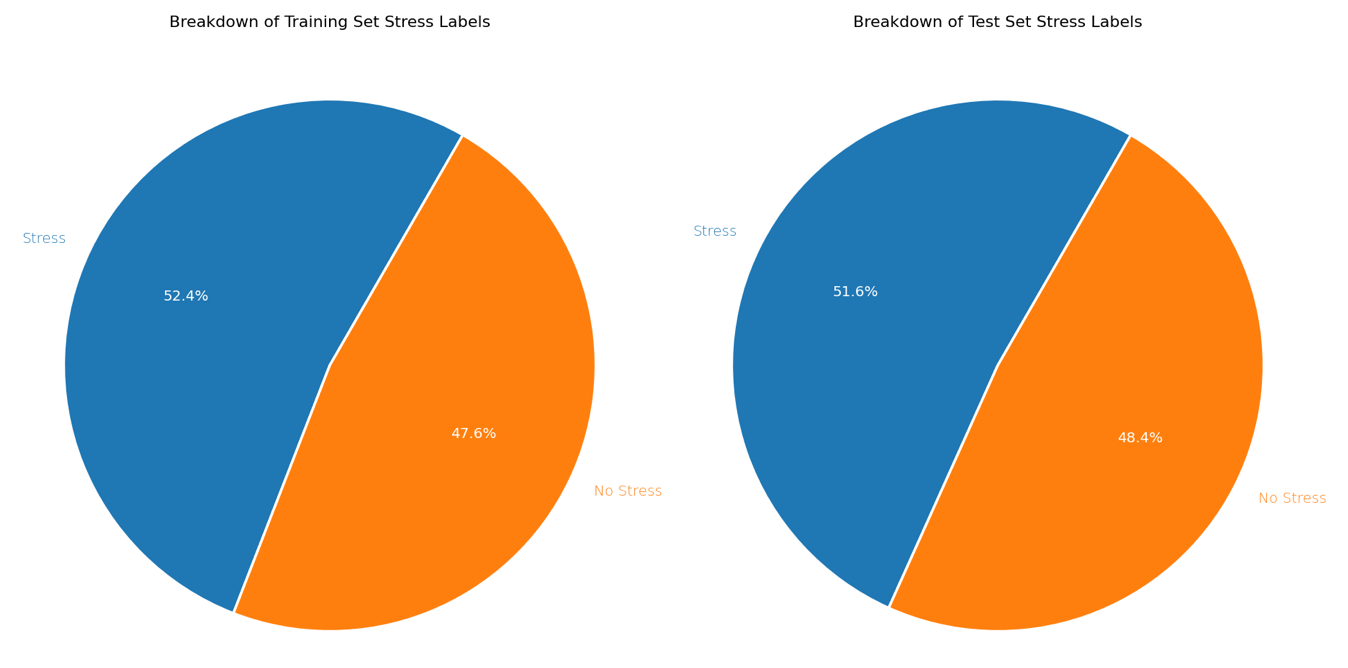}}
\caption{Training and test set stress label distribution.}
\label{fig:stress_label_distribution}
\end{figure}

\begin{figure}[!h]
\centering
{\includegraphics[width=0.83\columnwidth]{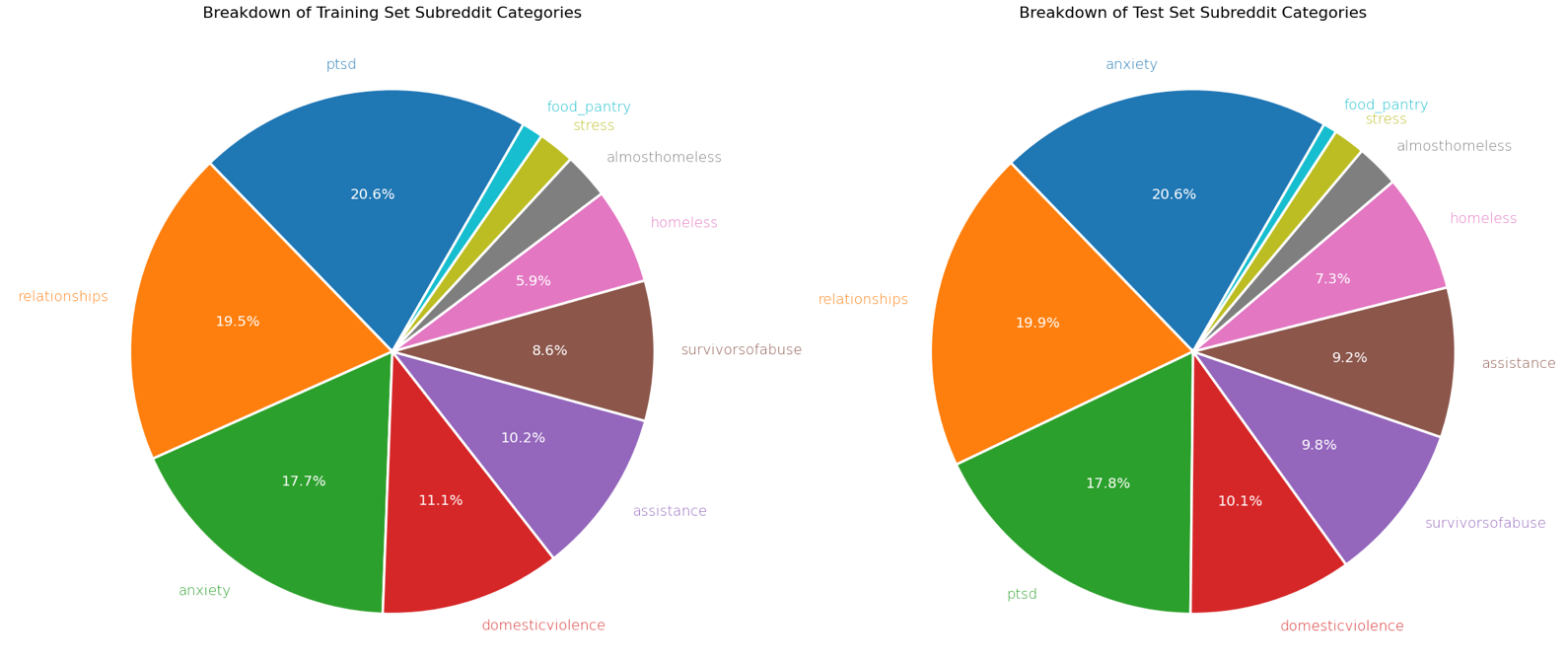}}
\caption{Training and test set subreddit label distribution.}
\label{fig:subreddit_distribution}
\end{figure}

In Figure~\ref{fig:stress_label_distribution} and Figure~\ref{fig:subreddit_distribution}, we show the data distribution of our stress and context (subreddit) labels.

\newpage 

\subsection{MentalRoBERTa}
MentalRoBERTa is a RoBERTa-based language model \cite{liu2019roberta} that was pre-trained on a corpus of 13.7M sentences from Reddit that were posted on mental health-related subreddits, including, but not limited to, ``r/Anxiety'' and ``r/Depression''. When training classifiers for stress and context classification tasks, we used the pre-trained MentalRoBERTa model on Hugging Face’s model repository, available at
\url{https://huggingface.co/mental}, and fine-tuned the model on the \Dreaddit dataset, using either the stress or context labels, for five epochs with a learning rate of 1e-4. 


\subsection{Qualitative Examples}
In Tables \ref{tab:results_explainability_qualitative_anxiety}, \ref{tab:results_explainability_qualitative_assistance}, and \ref{tab:results_explainability_qualitative_relationships}, we show qualitative examples of our MCTS method for explainability, with examples from each of three subreddits---``anxiety,'' ``assistance,'' and ``relationships''---from both the MLP and MentalRoBERTa\textsuperscript{FT} models.

\begin{table*}[h]
\small 
\centering
\scalebox{0.90}{
\begin{tabular}{c  c | p{0.7\textwidth} | c | c}
\toprule
\bf{Model} &  \bf{Category} & \bf{\quad \quad\quad\quad\quad \quad\quad\quad\quad \quad \quad\quad Text (subreddit = ``r/Anxiety'')} & \bf{Stress} & \bf{Entropy} \\ 
\toprule
 & \multirowcell{8}{\textbf{\textcolor{black}{Original}}} & Lately I’ve just been having that terrible feeling in the pit of my stomach and also a feeling of nausea like I constantly need to throw up. I’m sleeping normal but still feeling so tired and drained and can’t really focus at work and because of that I feel like my work performance is slipping up. I am constantly afraid that I’m going to lose my job and that my manager hates me. This has been happening so much more frequently. About a week ago my doc gave me prozac (once a day) and xanax (only as needed) prescriptions and I feel like it’s helped with the bigger attacks and some dark thoughts but now its almost like just a little constant anxiety all the time and it sucks. & \multirowcell{8}{1.000} & \multirowcell{8}{0.000} \\  
 \midrule
 \multirowcell{16}{\textbf{MLP}} & \multirowcell{8}{\textbf{\textcolor{orange}{Dependent}}} & \textcolor{lightgray}{Lately I’ve just been having that terrible feeling in \textcolor{orange}{the pit of my stomach and also a feeling of nausea like I constantly need to throw up. I’m sleeping normal but still feeling so tired and drained and can’t really focus} at work and because of that I feel like my work performance is slipping up. I am constantly \textcolor{orange}{afraid that I’m going to lose my job and that my manager hates me.} This has been happening so much more frequently. About a week ago my doc gave me prozac (once a day) and xanax (only as needed) prescriptions and \textcolor{orange}{I feel like it’s helped with the bigger attacks and some dark thoughts but now its almost like} just a little constant anxiety all the time and it sucks.} & \multirowcell{8}{0.933} & \multirowcell{8}{0.300} \\ \cdashlinelr{2-5}
 & \multirowcell{8}{\textbf{\textcolor{teal}{Independent}}} & \textcolor{lightgray}{Lately \textcolor{teal}{I’ve just been having that} terrible feeling in the pit of my \textcolor{teal}{stomach and also a feeling} of nausea like I constantly need to throw up. I’m sleeping normal but still feeling so tired and drained and can’t really focus at work and because of that I feel like my work performance is slipping up. I am constantly afraid that I’m going to lose my \textcolor{teal}{job and that my manager hates me. This has been happening so much more frequently. About a week ago} my doc gave me prozac (once a day) and xanax (only as needed) prescriptions and I feel like it’s helped with the bigger attacks and some dark thoughts but now its almost like just a little constant anxiety all the time and it sucks.} & \multirowcell{8}{0.489} & \multirowcell{8}{1.045} \\
\midrule
\multirowcell{16}{\textbf{Mental}\\\textbf{RoBERTa}\textsuperscript{FT}} & \multirowcell{8}{\textbf{\textcolor{orange}{Dependent}}} & \textcolor{lightgray}{Lately I’ve just been having that terrible feeling in \textcolor{orange}{the pit of my stomach and also a feeling of nausea like I constantly need to throw up. I’m sleeping normal but still feeling so tired and} drained and can’t really focus at work and because of that I feel like my work performance is slipping up. I am constantly \textcolor{orange}{afraid that I’m going to lose my job} and that my manager hates me. This has been happening so much more frequently. About a week ago my doc gave me prozac (once a day) and \textcolor{orange}{xanax (only as needed) prescriptions and I feel like it’s helped with the bigger attacks and some dark thoughts but now its almost like just} a little constant anxiety all the time and it sucks.} & \multirowcell{8}{0.998} & \multirowcell{8}{0.006} \\ \cdashlinelr{2-5}
 & \multirowcell{8}{\textbf{\textcolor{teal}{Independent}}} & \textcolor{lightgray}{Lately I’ve just been having that terrible feeling in the pit of \textcolor{teal}{my stomach and also} a feeling of nausea like I constantly need to throw up. I’m sleeping normal but still feeling so tired and drained \textcolor{teal}{and can’t really focus at work and because of that I feel like my work performance is slipping up. I am constantly afraid that I’m going to lose my job and that my manager hates me. This} has been happening so much more frequently. About a week ago my doc gave me prozac (once a day) and xanax (only as needed) prescriptions and I feel like it’s \textcolor{teal}{helped with the bigger attacks and} some dark thoughts but now its almost like just a little constant anxiety all the time and it sucks.} & \multirowcell{8}{0.670} & \multirowcell{8}{0.627}\\
\bottomrule
\end{tabular}
}
\caption{Qualitative examples from our MCTS explainability method for a post in the ``r/Anxiety'' subreddit. We show the full original text along with the context-dependent and context-independent explanations selected by MCTS using both the MLP and MentalRoBERTa\textsuperscript{FT} classifiers.}
\label{tab:results_explainability_qualitative_anxiety}
\end{table*}

\newpage
\begin{table*}[h]
\small 
\centering
\scalebox{0.90}{
\begin{tabular}{c  c | p{0.7\textwidth} | c | c}
\toprule
\bf{Model} &  \bf{Category} & \bf{\quad \quad\quad\quad\quad \quad\quad\quad\quad \quad \quad\quad Text (subreddit = ``r/Assistance'')} & \bf{Stress} & \bf{Entropy} \\ 
\toprule
 & \multirowcell{5}{\textbf{\textcolor{black}{Original}}} & I can’t ask my family because they don’t have the kind of money to help me. If anyone can help me even just a little bit, I would be ridiculously grateful. I just can’t even express what this has done to us. Yes, the bills are paid, but now we’re so anxious that we barely leave the house due to panic attacks. I’ve done things like ubereats but \$15 here and there isn’t even making a dent in what I need. & \multirowcell{5}{0.995} & \multirowcell{5}{0.616} \\  
 \midrule
 \multirowcell{10}{\textbf{MLP}} & \multirowcell{5}{\textbf{\textcolor{orange}{Dependent}}} & \textcolor{lightgray}{I can’t ask my family because \textcolor{orange}{they don’t have the kind of money to help me. If anyone can help me even} just a little bit, I would be ridiculously grateful. I just can’t even express \textcolor{orange}{what this has done to} us. Yes, the bills are paid, but now we’re so anxious that we barely leave \textcolor{orange}{the house due to panic attacks. I’ve done things like ubereats but \$15 here and there isn’t even making} a dent in what I need} & \multirowcell{5}{0.723} & \multirowcell{5}{0.640} \\ \cdashlinelr{2-5}
 & \multirowcell{5}{\textbf{\textcolor{teal}{Independent}}} & \textcolor{lightgray}{I can’t ask my family because \textcolor{teal}{they don’t have the kind} of money to help me. If anyone can help \textcolor{teal}{me even just a little} bit, I would be ridiculously grateful. I just can’t even express what this has done to us. Yes, the bills are paid, but now \textcolor{teal}{we’re so anxious that we barely leave the house due to} panic attacks. I’ve done things like ubereats but \$15 here and there isn’t even making a dent in what I need.} & \multirowcell{5}{0.584} & \multirowcell{5}{1.064} \\
\midrule
\multirowcell{10}{\textbf{Mental}\\\textbf{RoBERTa}\textsuperscript{FT}} & \multirowcell{5}{\textbf{\textcolor{orange}{Dependent}}} & \textcolor{lightgray}{I can’t ask my family because they don’t have the kind of \textcolor{orange}{money to help me. If anyone can help me even just a little bit, I would be ridiculously grateful. I just can’t even express what this has} done to us. Yes, the bills are paid, but now \textcolor{orange}{we’re so anxious that we} barely leave the house due \textcolor{orange}{to panic attacks. I’ve done things like} ubereats but \$15 here and there isn’t even making a dent in what I need.} & \multirowcell{5}{0.999} & \multirowcell{5}{0.005} \\ \cdashlinelr{2-5}
 & \multirowcell{5}{\textbf{\textcolor{teal}{Independent}}} & \textcolor{lightgray}{I can’t ask my family because they don’t have the kind \textcolor{teal}{of money to help me}. If anyone can help me even \textcolor{teal}{just a little bit, I} would be ridiculously grateful. I just can’t even express what this has done to us. Yes, the \textcolor{teal}{bills are paid, but now we’re so anxious that we barely leave the house due to panic attacks. I’ve done things like ubereats but \$15 here} and there isn’t even making a dent in what I need.}  & \multirowcell{5}{0.478} & \multirowcell{5}{0.518} \\
\bottomrule
\end{tabular}
}
\caption{Qualitative examples from our MCTS explainability method for a post in the ``r/Assistance'' subreddit. We show the full original text along with the context-dependent and context-independent explanations selected by MCTS using both the MLP and MentalRoBERTa\textsuperscript{FT} classifiers.}
\label{tab:results_explainability_qualitative_assistance}
\end{table*}


\begin{table*}[h]
\small 
\centering
\scalebox{0.90}{
\begin{tabular}{c  c | p{0.7\textwidth} | c | c}
\toprule
\bf{Model} &  \bf{Category} & \bf{\quad \quad\quad\quad\quad \quad\quad\quad\quad \quad \quad\quad Text (subreddit = ``r/Relationships'')} & \bf{Stress} & \bf{Entropy} \\ 
\toprule
 & \multirowcell{8}{\textbf{\textcolor{black}{Original}}} & We seem to be talking and accidentally being together more often in school, making what I think are feelings towards her only stronger. I can't bring myself to bring this up with her because I'm scared that we will have a repeat of February again. I love her so much but I feel that if I have these feelings about other girls am I really devoted to her? This is in no way her fault, she has done nothing to deserve my questioning of my decision, this is my problem and mine alone. I am reluctant to bring this up with her because I'm worried that she might break up with me because I do truly still love her I'm just wondering if this other girl is a passing thought more focused than earlier and something I can overcome. & \multirowcell{8}{0.999} & \multirowcell{8}{0.000} \\  
 \midrule
 \multirowcell{16}{\textbf{MLP}} & \multirowcell{8}{\textbf{\textcolor{orange}{Dependent}}} & \textcolor{orange}{We seem to be talking} \textcolor{lightgray}{and accidentally being together more often in school, making what I think are feelings towards her only stronger. I can't} \textcolor{orange}{bring myself to bring this up with her because I'm scared that we will have a repeat of February again. I love her so much but I feel that if I have these feelings about other girls am} \textcolor{lightgray}{I really devoted to her? This is in no way her fault, she has done nothing to deserve my questioning of my decision, this is my problem and mine alone. I} \textcolor{orange}{am reluctant to bring this up with her because I'm worried that she might break up with me because I do truly still love her}  \textcolor{lightgray}{I'm just wondering if this other girl is a passing thought more focused than earlier and something I can overcome.} & \multirowcell{8}{0.734} & \multirowcell{8}{0.437} \\ \cdashlinelr{2-5}
 & \multirowcell{8}{\textbf{\textcolor{teal}{Independent}}} & \textcolor{lightgray}{We} \textcolor{teal}{seem to be talking and} \textcolor{lightgray}{accidentally being together more often in school, making}  \textcolor{teal}{what I think are feelings} \textcolor{lightgray}{towards her only stronger. I can't bring myself to bring this up with her because I'm scared that we will have a repeat of February again. I love her so much but I feel that if I have these feelings about other girls am I really devoted to her? This is} \textcolor{teal}{in no way her fault, she has done nothing to deserve my questioning of my decision, this}  \textcolor{lightgray}{is my problem and mine alone. I am reluctant to bring this up with her because I'm worried that she might break up with me because I do truly still love her I'm just wondering if this other girl is a passing thought more focused than earlier and something I can overcome.} & \multirowcell{8}{0.510} & \multirowcell{8}{1.043} \\
\midrule
\multirowcell{16}{\textbf{Mental}\\\textbf{RoBERTa}\textsuperscript{FT}} & \multirowcell{8}{\textbf{\textcolor{orange}{Dependent}}} & \textcolor{lightgray}{We seem to be talking and accidentally being together more often \textcolor{orange}{in school, making what I think are feelings towards her only stronger. I can't bring myself to bring this up} with her because I'm scared that we will have a repeat of February again. I love her so much but I feel that if \textcolor{orange}{I have these feelings about other girls am I really devoted to her? This is in no way her fault, she has done nothing} to deserve my questioning of my decision, this is my problem and mine alone. I am reluctant to bring this up with her because I'm worried \textcolor{orange}{that she might break up with me because I do truly still love her I'm just} wondering if this other girl is a passing thought more focused than earlier and something I can overcome.} & \multirowcell{8}{0.998} & \multirowcell{8}{0.030} \\ \cdashlinelr{2-5}
 & \multirowcell{8}{\textbf{\textcolor{teal}{Independent}}} & \textcolor{lightgray}{We seem to be talking and accidentally being together more often in school, making what I think are feelings towards \textcolor{teal}{her only stronger. I can't bring myself to bring this up with her because I'm scared that we will have a repeat of February again. I love her so much but I feel that if I have these feelings about other girls am} I really devoted to her? This is in no way her fault, she has done nothing to deserve my questioning of my decision, this is my problem and mine alone. I am \textcolor{teal}{reluctant to bring this up} with her because I'm worried that she might break up with me because I do truly still love her I'm just wondering if this other \textcolor{teal}{girl is a passing thought} more focused than earlier and something I can overcome.} & \multirowcell{8}{0.712} & \multirowcell{8}{0.444} \\
\bottomrule
\end{tabular}
}
\caption{Qualitative examples from our MCTS explainability method for a post in the ``r/Relationships'' subreddit. We show the full original text along with the context-dependent and context-independent explanations selected by MCTS using both the MLP and MentalRoBERTa\textsuperscript{FT} classifiers.}
\label{tab:results_explainability_qualitative_relationships}
\end{table*}

\end{document}